\documentclass[review,12pt,numafflabel]{elsarticle}
\usepackage{booktabs}
\usepackage{longtable}
\usepackage{xurl}

\usepackage{array}
\usepackage{graphicx}
\usepackage{amsmath}
\usepackage{amssymb}
\usepackage{pifont}

\newcommand{\cmark}{\ding{51}}
\newcommand{\xmark}{\ding{55}}

\journal{Information and Software Technology}

\begin{document}

\begin{frontmatter}

\title{Real Computation or Merely Naming? A Pre-Registered Mechanism Audit
of 25 Cognitive-Architecture Modules in an LLM-Based Agent System}

\author[1]{Peter Chen\fnref{fn1}}
\author[1]{Ni Zhang}
\author[1]{Ying Xu\fnref{fn1}}
\author[1]{Siyuan Zhang\corref{cor1}}
\fntext[fn1]{These authors contributed equally to this work.}
\cortext[cor1]{Corresponding author.}
\ead{lelandsy28@163.com}
\affiliation[1]{organization={Shenyang Institute of Technology},
city={Shenyang}, country={China}}

\begin{abstract}
\textbf{Context:} Agent systems built on local large language models are
increasingly marketed through the vocabulary of cognitive architecture:
modules named \emph{global workspace}, \emph{theory of mind},
\emph{causal reasoning}, or \emph{active inference} are listed as
features, and the presence of the name is taken as the presence of the
mechanism.
\textbf{Objectives:} We ask the prior question -- does the named
computation actually happen -- and contribute a transferable
software-verification method for the claim--implementation gap: a
four-question static audit protocol with line-number evidence and
call-chain tracing, applicable in principle to any component whose
documentation speaks in mechanism names and demonstrated here on two
systems.
\textbf{Methods:} We audit a system of this class whose
documentation claims 25 cognitive mechanisms, pre-register falsifiable
predictions for 11 modules and execute the corresponding interventions,
replicate the protocol on the IIT-$\phi$ core of a second,
independent system that pre-commits to its own falsification criteria
(one failure class, one module deep), and run two repair experiments as
positive controls.
\textbf{Results:} In 18 of 25 modules the computed value never reaches
any production output; seven cross-cutting findings include three
idle-wiring shapes new to the dead-code and code-smell literatures we
survey
(\emph{connected-but-never-called}, \emph{called-but-inert},
\emph{called-but-content-blind}). Of 61 frozen criteria across 12
interventions, 53 were confirmed, two partially adjudicated, five
falsified, and one adjudication-ambiguous. The repairs raised planner
task success from 0.028 to 1.000 and cut state-controller error by
52\%.
\textbf{Conclusion:} Module names and module computations must be
audited as independent claims. The protocol and the five-class failure
taxonomy are reusable as a verification method for the
claim--implementation gap, demonstrated at two-system, case-level
depth; the three idle-wiring shapes are architecture-defect patterns
that reach beyond cognitive-architecture vocabulary.
\end{abstract}

\begin{keyword}
mechanism audit \sep cognitive architecture \sep pre-registration \sep
negative results \sep global workspace theory \sep LLM agents \sep
static analysis \sep artifact evaluation
\end{keyword}

\end{frontmatter}

\section{Introduction}
Agent systems built on local large language models are increasingly
described using the vocabulary of cognitive science. The substrate is
real: code-generating models \cite{chen2021codex} plugged into
tool-use, reflection, and role-playing loops
\cite{react2023,shinn2023,madaan2023,park2023generative,wang2024agentsurvey}
produce systems whose documented failure modes, including
hallucination \cite{ji2023}, are as well known as their capabilities.
A repository announces
a \emph{global workspace} for consciousness, a \emph{theory-of-mind}
module, a \emph{dopamine system} for reward prediction, a \emph{causal
reasoning} engine, and a \emph{least-action planner}. The rhetorical force
of these names is considerable: they suggest that decades of cognitive
science have been operationalized in software. The engineering reality,
however, is rarely examined. A name can be attached to a keyword counter,
a constant, or to code that is never called; the claim and the computation
are logically independent, and only the second can do work. Even
self-correction machinery, one of the field's standard answers to LLM
unreliability, does not by itself repair reasoning errors
\cite{huang2024}.

This paper is an audit of that gap. The audit doubles as a
software-verification method for the claim--implementation gap -- four
checkable questions per module, line-number evidence, call-chain and
value-to-output tracing, a static complement to test-based verification
\cite{ammann2017testing} -- and its failure shapes, modules that are called
yet contribute nothing to the output, are architecture-defect patterns
relevant to dead-code analysis and integration testing, not facts about
one repository. Our object system is a member of this
class: an open repository that couples a shell of 25 named
cognitive modules to an LLM-based kernel, with documentation that presents
the mechanism names as capabilities (its own position paper lists
``causal reasoning expert'' among its core innovations). We make no
statistical sampling claim: the audit treats the repository as an
instance of the class, and transfer is probed by the cross-system check
of Section~\ref{sec:aura}. We ask, for each
module, a question that is mechanically checkable: \emph{is there real
computation behind the name, and does its value reach the output?}

Two prior studies in this line developed and applied the methodology. A companion
audit of an emergence-marker battery \cite{ourpaper7} showed that
``emergence'' claims in agent systems can be falsified by
random-autonomy controls; an audit of a neurotransmitter-style modulator
\cite{ourpaper11} showed that a single einsum-level defect can render an
entire claimed mechanism inert while the code remains importable and
executable. The present study generalizes from single mechanisms to a
whole architecture, and adds two elements neither prior study had:
a \emph{system-scale taxonomy} of failure classes, and \emph{positive
controls} -- modules that fail the static audit but succeed after repair,
demonstrating that the audit distinguishes broken implementations from
unimplementable claims.

Our contribution is deliberately not the discovery of a surprising
phenomenon: most findings here are visible from reading code. The
contribution is quantification, pre-registration, and a reusable
protocol -- the same three values that motivated the companion audits
\cite{ourpaper7,ourpaper11}: (i) a four-question static audit applied to
all 25 modules with line-number evidence; (ii) 12 pre-registered
intervention experiments with 61 frozen criteria, executed before any
intervention data were collected; (iii) a five-class taxonomy of
mechanism failure with distinct repair prospects, supported by two
repair experiments and by one cross-system replication (one class, one
module deep) on the IIT core of a second architecture; and (iv) seven
cross-cutting findings about how
naming decouples from computation in this system class, including three
idle-wiring shapes that, to our knowledge, have not been catalogued in
the dead-code and code-smell literature we survey.

\section{Related Work}
\textbf{Audits of specific mechanisms.} The emergence-marker battery audit
\cite{ourpaper7} tested four ``emergence verification'' claims with
pre-registered controls and found none survived a random-autonomy
baseline. The modulator audit \cite{ourpaper11} pre-registered a causal
test of a neurotransmitter-style attention modulator and located a
single defect that severed all influence on the output. Both studies
argued that the value of such audits lies in rigor and protocol, not in
the surprise of the findings; we adopt that position and scale it from
one mechanism to an architecture.

\textbf{Concurrent evidence audits.} Concurrent work reports that
AI-scientist agents often answer benchmark questions without using the
supplied evidence \cite{zhang2026measuring}, and that LLM-generated
kernels which pass numerical checks can fail under scale shift
\cite{suwandi2026kernaut}. Those studies audit evaluation methodology
and artifact robustness at the level of agent runs; the present audit
performs causal interventions on the mechanisms of one specific
architecture, with positive controls and cross-system validation, and
asks whether the claimed machinery ever reaches the output at all.
Assessment itself can also inflate the phenomenon under study: reported
emergent capabilities can be artifacts of the metric rather than
properties of the model \cite{schaeffer2023}, which is why the present
audit grounds each claim in code that either exists or does not.

\textbf{Harness-performance measurement.} Two concurrent benchmark
efforts evaluate execution harnesses at the performance layer: a
five-measure framework attributing benchmark scores between models and
harnesses across 52{,}168 trajectories \cite{zhang2026whenagents}, and a
seven-component reliability benchmark with run-level failure attribution
across 3{,}420 runs \cite{cao2026tracetrial}. Both ask how well the
machinery performs once it exists; the present audit asks whether the
named machinery exists at all and whether its computed value ever
reaches an output (Section~\ref{sec:static}). The layer difference has
five concrete consequences. (i)~Existence versus performance: a
called-but-inert module executes on every trajectory, spends tokens on
every run, and is scored by task-level metrics as working
infrastructure, yet contributes nothing to the answer; establishing the
absence of causal effect at component granularity requires tracing the
code path itself, without a trajectory corpus. (ii)~A defect vocabulary
that performance
measurement does not possess: the three idle-wiring shapes
(connected-but-never-called, called-but-inert, called-but-content-blind)
and the four-question protocol that surfaces them
(Section~\ref{sec:protocol}) are addressed to module-internal wiring,
which run-level scoring does not inspect. (iii)~A different unit of
analysis: the audited objects are module-level mechanism claims carried
by cognitive-architecture vocabulary (global workspace, theory of mind,
active inference), not run-level harness components; to our knowledge,
no audit of this class exists.
(iv)~Falsification structure: 61 frozen criteria with five falsified and
reported as such, a second system replicating the protocol
under its own
pre-committed criteria (Section~\ref{sec:aura}), and repair experiments
as positive controls; as presented, the two benchmarks center run-level
scoring rather than pre-registered mechanism falsification, and the two
structures are complementary. (v)~Attribution semantics: their failure
attribution assigns a
failed run to a component; ours assigns causal contribution to a module
and separates naming from computation.

\textbf{Global workspace theory and its imitations.} Global workspace
theory \cite{baars1988,baarsfranklin2003} posits a broadcast architecture
in which module contents compete for access to a limited-capacity
workspace whose winning content is broadcast to all consumers. The theory
makes functional commitments -- content sensitivity, competition,
broadcast -- that are testable in software. Systems that implement only
the name (a salience table plus an argmax) inherit none of these
commitments; Section~\ref{sec:interventions} shows what a pre-registered
test of those commitments looks like.

\textbf{Reproducibility and artifact evaluation.} The recurring failure
to reproduce published systems research
\cite{collberg2016,lipton2019} and the credibility crisis it exposed
\cite{ioannidis2005,fanelli2012} motivated preregistration reform
\cite{nosek2018,chambers2022}, leakage-aware reproducibility standards
\cite{kapoor2023leakage}, artifact badging, and
reproducibility checklists \cite{acm2020,pineau2021}. Those processes
ask whether \emph{claimed results} recompute; the present audit asks a
prior question, whether \emph{claimed mechanisms} compute at all. The two
are complementary: a system can pass every results-level check while
none of its advertised cognitive mechanisms influences its behavior.

\textbf{Dead-code and integration analysis.} Static analysis for dead
code is classical: program slicing traces the statements that affect a
variable's value \cite{weiser1984slicing}, compilers eliminate
provably unreachable code, and code-smell catalogs flag speculative
generality -- abstractions built for anticipated uses that never
arrive \cite{fowler1999refactoring}. The failure modes catalogued here
are adjacent but distinct. Several audited mechanisms are alive in every
syntactic sense -- imported, called, executed -- yet fail to influence
output because their result is discarded, their inputs are constants, or
their wiring is unidirectional into a write-only sink. Of the three
idle-wiring shapes, connected-but-never-called is the closest to
classical dead code (its sole caller is itself never imported, finding 2
below); called-but-inert involves code that executes on every run and is
reachable by construction, so reachability analysis and profiling both
pass it; and called-but-content-blind requires checking \emph{which}
inputs a called function reads, a data-flow question one level deeper
than slicing for a designated variable. None of the three is captured by
smell catalogs, which address design quality rather than value arrival.
These distinctions are why our protocol tracks \emph{value-to-output}
rather than mere execution; we do not claim the shapes are unnamable in
principle, only that they are not catalogued in the literatures we
survey.

\section{Object System and Audit Protocol}
\label{sec:protocol}
\subsection{Object}
The object system couples 25 named cognitive modules (directories
\texttt{fusion/}, \texttt{core/cognitive/}, \texttt{emotion/},
\texttt{planning/}, \texttt{s9\_morph/}, among others) to an LLM-based
kernel. Its documentation lists mechanisms from cognitive science --
global workspace, theory of mind, causal reasoning, active inference,
dopamine reward prediction, holographic memory, least-action planning --
as architectural features. We audit the repository as shipped; we do not
evaluate the LLM kernel itself.

\subsection{Four-question static audit}
Four read-only exploration agents audited all 25 modules in parallel
(traversing the repository and its call chains with \texttt{grep},
virtual environment excluded). The agents were LLM-driven read-only
sessions with a fixed four-question prompt; they had no write access to
the repository and reported one verdict per module. Every verdict is
anchored to line numbers archived with the audit log, so each Table~1
entry can be re-checked independently against the named revision. We did
not run a separate inter-rater reliability pass over the agent verdicts;
the 11-module intervention layer (Section~\ref{sec:interventions})
provided an independent dynamic check on that subset, and we flag
auditor-reliability measurement as a limitation (Section~\ref{sec:aura}
and Section~7.2). For each module they answered:
\begin{enumerate}
\item \textbf{What is the core computation?} (evidence by line number)
\item \textbf{How should it be classified?} (real algorithm; rule table;
renamed heuristic; stub; name without definition)
\item \textbf{Does the value reach any output?} (tracing production call
paths; strict criterion: influences control flow or returned content)
\item \textbf{What are the side effects?}
\end{enumerate}
The strict reachability criterion deserves emphasis. A module whose
output is written into a log field or a display string is credited with
arrival only if that string can change behavior; decorative text
prefixes and inert metadata do not count.

\subsection{Methodological disclaimer}
Most verdicts in Section~\ref{sec:static} are visible from reading code.
The incremental value of the intervention experiments
(Section~\ref{sec:interventions}) is not surprise but quantification,
reproducibility, and pre-registration -- the same position taken by the
modulator audit \cite{ourpaper11}, whose decisive defect was likewise
visible statically. One module (the nine-state controller) had a
genuinely open outcome and was ranked first for experimentation for
exactly that reason.

\section{Static Audit Results}
\label{sec:static}
\subsection{Module-by-module verdicts}
Table~\ref{tab:modules} summarizes the audit. Of 25 modules, 18 fail
outright (computed value never reaches production output). The remaining
7 are flagged partial: 4 reach output only as decorative text or inert
metadata (modules 6, 16, 17, 20), 2 ship real computation that is then
discarded or mocked (modules 14, 25), and 1 terminates in a live LLM call
while computing seven sub-results that are never read (module 18). None
of the 7 influences control flow.

\begingroup
\footnotesize
\begin{longtable}{@{}>{\raggedright\arraybackslash}p{0.4cm}>{\raggedright\arraybackslash}p{4.3cm}>{\raggedright\arraybackslash}p{2.0cm}>{\raggedright\arraybackslash}p{1.1cm}>{\raggedright\arraybackslash}p{4.0cm}@{}}
\caption{Static audit of all 25 modules. \xmark: value never reaches any
production output (18 modules). \checkmark$^{a}$: reaches output only as
decorative text or inert metadata (4 modules, never influencing control
flow). $\oslash$: partial -- real computation that is then discarded
(module 14), a live LLM call with never-read sub-results (module 18), or
mocked success (module 25). Suture identifiers (``see sutureNN'') point
forward to the intervention scorecard (Table~\ref{tab:scorecard}); the
numbering is not contiguous because identifiers are stable keys in the
audit log's intervention registry. The Class column records the
auditor's per-module label; Section~\ref{sec:taxonomy} maps the
interventions into five failure classes. Line numbers refer to the
audited repository revision, whose content hash and date are archived
with the audit log.}
\label{tab:modules} \\
\toprule
\# & Module (claimed mechanism) & Class & Reach & Key evidence (line numbers in repository) \\
\midrule
1 & \texttt{causal\_reasoning} (causal reasoning) & name$\neq$def & \xmark &
hardcoded strength table (L269--275); \texttt{simulate\_intervention}
returns \{\} (L482--485); direction set by correlation sign alone (L530--534) \\
2 & \texttt{theory\_of\_mind} & rule table & \xmark & beliefs keep last
write only (L106); \texttt{mentalize} returns $0.7 \cdot 0.9^{\mathrm{depth}}$ (L291) \\
3 & \texttt{world\_model} & rule table & \xmark & ``state similarity''
compares entity counts (L637--642); dynamics reduce to $v\,\Delta t$ (L726) \\
4 & \texttt{self\_awareness} & rule table & \xmark & hallucination =
keyword count (L238--269); confirmation bias computes then
\texttt{return 0.0} (L322); uncertainty = \texttt{random.choice} (L514) \\
5 & \texttt{active\_inference} (FEP) & name$\neq$def & \xmark & real
binary Shannon entropy (L207) but no argmin/gradient; see suture23 \\
6 & \texttt{negative\_entropy} & rule table & \checkmark$^{a}$ &
keyword match + fixed coefficients; sign of ``negentropy'' decided by a
coin flip (\texttt{random.random()}, L332) \\
7 & \texttt{negative\_entropy\_\allowbreak learning} & rule table & \xmark &
template strings; hardcoded \texttt{confidence=0.7} (L141); injected LLM
adapter never used \\
8 & \texttt{intrinsic\_motivation} & rule table & \xmark & handwritten
product (L110) + noise-argmax via \texttt{random.uniform} (L134) \\
9 & \texttt{emotional\_intelligence} & rule table & \xmark & keyword
counts (L210--213) + fixed bias table (L267--304) \\
10 & \texttt{holographic\_memory} (HRR) & stub & \xmark & encoding =
\texttt{md5} $\to$ \texttt{random.seed} $\to$ \texttt{random.random()}
(L96--99); no convolution, no binding; see suture22 \\
11 & \texttt{potential\_observer} & rule table & \xmark & fixed-weight
dot product (L338--341); a real \texttt{lstsq} learner exists but is
never called (L288) \\
12 & \texttt{least\_action\_planner} & pipeline defect & \xmark & A*
search is real; ``action'' term does not drive the search (L191--202);
repairable, see suture29 \\
13 & \texttt{dopamine\_system} (RPE) & divergent algorithm & \xmark & TD
learner diverges on the 5-state MDP ($\max|V|\sim 10^{131}$; up to
$6.6\times 10^{232}$ on the 7-state, Table~\ref{tab:scorecard}); reward
value discarded by the production caller (\texttt{civilization\_os.py:\allowbreak 369}); see suture20 \\
14 & \texttt{nine\_state\_controller} & real algorithm, dropped & $\oslash$ &
trainable MLP is real; transition matrix and gates computed then
discarded (L312--319, L341--400); see suture19 \\
15 & \texttt{nine\_state\_core} & rule table & \xmark & hardcoded trait
argmax; ``nine states'' are 11 enum members \\
16 & \texttt{seven\_emotions} & decorative & \checkmark$^{a}$ & keyword
+ fixed increments; output is a ``\texttt{[FIRM]}'' text prefix fed a
constant \texttt{\{"success": True\}}; ``seven emotions'' are 12 \\
17 & \texttt{intuition\_engine} & weak & \checkmark$^{a}$ & real nearest
neighbor (L98--102), but see suture24: reachable set is empty \\
18 & \texttt{brain\_layer\_integrator} & LLM wrapper & $\oslash$ &
15-stage pipeline terminates in one LLM call (L1278--1288); seven
sub-results computed and never read \\
19 & \texttt{global\_workspace} (GWT) & name$\neq$def & \xmark &
hardcoded salience table 0.8/0.7/0.6/0.5 (L153--251); see suture25 \\
20 & \texttt{mirror\_neurons} & weak & \checkmark$^{a}$ & keyword hit
$\times$ weights + canned phrases; copy-paste bug \texttt{'!!!' in text
or '!!!' in text} (L244) \\
21 & \texttt{neuro\_modulation} & real algorithm, unwired & \xmark &
real weighted sum + normalization (L274--308); hand-fixed coefficients;
modulation has no production injection point \\
22 & \texttt{thalamic\_filter} & dead wiring & \xmark & regex + static
routing table (L262); see suture26 \\
23 & \texttt{hemisphere\_coordination} & stub & \xmark & \texttt{random.uniform}
as originality (L355); \texttt{md5} as features (L413); word reversal as
association (L390); see suture27 \\
24 & \texttt{attention\_competition} & real algorithm, unwired & \xmark &
real proportional allocation (L277--286); see suture28 \\
25 & \texttt{cognitive\_kernel} (root) & rule table & $\oslash$ &
$\sim$30 regexes $\to$ shell command templates; without a world adapter
every action is mocked as success (L1238) \\
\bottomrule
\end{longtable}
\endgroup

\subsection{Cross-cutting findings}
\label{sec:crosscutting}
Seven findings span modules and are, in our judgment, more informative
than any single verdict:
\begin{enumerate}
\item \textbf{Almost nothing reaches the output.} Only 4 of 25 modules
reach production output in any form (as lazy metadata, canned empathetic
phrases, prompt text, or a text prefix), and none influences control flow.
The one live production call, the 15-stage integrator's LLM request
(module 18), carries none of its seven computed sub-results, which is why
its audit verdict is partial rather than arrival.
\item \textbf{Dead call chains are pervasive.} The sole caller of two
audited modules is itself never imported; three factory functions have no
production callers; a controller calls a method and an enum member that
do not exist.
\item \textbf{Hardcoded constants impersonate measurements.} A constant
\texttt{\{"success": True\}} is fed to two modules; salience
(0.8/0.7/0.6/0.5), initial correlations (0.5), and confidences
(0.9/0.8/0.7) are fixed literals; random Gaussian vectors stand in for
learned embeddings when an optional dependency is absent.
\item \textbf{Names disagree with their own counts.} ``Seven emotions''
are 12; ``nine states'' are 11 -- the naming layer does not even parse
itself.
\item \textbf{Documentation references code that does not exist.} Three
documented module paths are absent; one package imports a nonexistent
submodule, raising \texttt{ModuleNotFoundError} at import.
\item \textbf{One genuine cross-module side effect.} The holographic
memory encoder calls \texttt{random.seed(...)}, reseeding the process
global RNG and thereby perturbing every other consumer of \texttt{random}
in the process, including the coin flip in module~6
(\texttt{negative\_entropy}) and the target template selection of
module~5 (\texttt{active\_inference}) (confirmed experimentally, suture22).
\item \textbf{The wiring itself idles, in three new shapes.} Beyond
``computed then discarded,'' the audit found: \emph{connected-but-never-called}
(the thalamic filter connects four ``deep fusion'' modules and calls none
of them); \emph{called-but-inert} (the hemispheric broadcast method is
literally \texttt{pass}; five different modulator values map to the same
balance output); and \emph{called-but-content-blind} (the global
workspace's consciousness score never reads module content). These three
shapes are invisible to execution-based profiling -- everything runs --
and to reachability analysis -- everything is connected.
\end{enumerate}

\subsection{Ranking by falsifiability and audit cost}
To select intervention targets we scored the seven most promising modules
on F1 (does a quantitative prediction exist for pre-registration?), F2
(running gain: can execution provide information beyond reading code?),
F3 (impact: is the module a headline claim?), and C (cost, reversed;
pure-CPU minute-scale experiments score highest). The top three spanned
three different suspected failure classes: the nine-state controller
(open outcome: the classifier and training loop are real), the dopamine
system, and the causal reasoning engine. The remaining eight
primary-system targets were frozen after the static audit and before
any intervention data, each with its own \texttt{PREREGISTRATION.md} in
the audit log, covering the remaining suspected failure classes and the
headlined dead-wiring cases of Table~\ref{tab:modules}.
Section~\ref{sec:interventions}
reports all interventions; the complete scoring table is in the
supplementary audit log.

\section{Pre-Registered Intervention Experiments}
\label{sec:interventions}
Each intervention followed the same discipline: a
\texttt{PREREGISTRATION.md} with frozen, quantitative criteria was
committed before any intervention data were collected; results files
record failures with the same prominence as confirmations; analysis is
script-generated and re-runnable in minutes on CPU. Twelve interventions
were executed (sutures 19--29 on the primary system, suture38 on the
cross-system target of Section~\ref{sec:aura}) -- eleven primary-system
modules plus the second system's IIT core. Two modules yielded
positive controls after repair (Section~\ref{sec:repairs}).

\subsection{Scorecard}
Table~\ref{tab:scorecard} summarizes 61 frozen criteria across the 12
interventions. Fifty-three were confirmed, two were partially adjudicated,
and five were falsified; one further verdict (suture19 P4) is
adjudication-ambiguous and discussed in a footnote. Four of the five
falsifications resolve in directions favorable to the systems under test
(two repairs that succeeded beyond prediction; two failures of the
second system's live path that its own null machinery correctly
adjudicates), which we report with the same prominence as confirmations.

\begingroup
\small
\begin{longtable}{@{}>{\raggedright\arraybackslash}p{1.15cm}>{\raggedright\arraybackslash}p{3.2cm}>{\raggedright\arraybackslash}p{1.35cm}>{\raggedright\arraybackslash}p{6.1cm}@{}}
\caption{Intervention scorecard. $\oslash$ = partially adjudicated.
$^{*}$Adjudication note for suture19 P4: the pre-registration defined
P4 as ``random-state significantly worse than broken $\Leftrightarrow$
the state carries information,'' with the falsified outcome upgrading
the conclusion from ``mis-wired'' to ``mechanism useless.'' The measured
outcome (random $\equiv$ broken) matches the falsified branch, and the
results file records ``P4 confirmed'' in the sense that the
broken-residual-carries-no-information claim is thereby established. We
report the verdict as ambiguous under the frozen wording rather than
retroactively rewording it.}
\label{tab:scorecard} \\
\toprule
Suture & Module & Criteria & Headline result \\
\midrule
19 & nine-state controller & 5: 3\cmark, 1\xmark, 1$^{*}$ &
Transition/gating perturbation leaves output bit-identical (dead code
confirmed); recognizer learns its task (acc 1.000 vs.\ chance 0.111);
the \texttt{broken} residual is information-free
(\texttt{broken} $\equiv$ \texttt{random}, identical MSE to 4 decimals).
\textbf{P3 falsified in the system's favor}: once wired with multiplicative
gating, error drops 52\% (Section~\ref{sec:repairs}). $^{*}$ \\
20 & dopamine system & 5: 3\cmark, 1$\oslash$, 1\xmark &
TD learner diverges ($\max|V|$ up to $6.6\times 10^{232}$ on the 7-state
MDP, $2.9\times 10^{131}$ on the 5-state; production $1.67\times 10^{10}$):
the update applies to the arrival state with
coefficient $1+\alpha e\gamma \approx 1.40 > 1$ (the product of the
TD learner's rate, trace, and discount terms). After repairing credit
assignment the learner converges (corr 0.99) but still underperforms a
textbook TD($\lambda$) by 0.32--0.70; and on the production state space
$V$ converges to the constant $E[r]/(1-\gamma)$ -- the state is a result
label, uninformative about the next step. \\
21 & causal reasoning & 6: 6\cmark &
Direction recovery accuracy 0.5000 on balanced synthetic graphs (chance);
direction is a pure function of the correlation sign; confounded
structures produce direct-edge false positives 3/3 with confidence
0.91--0.93; interventions propagate nothing (\texttt{downstream\_effects}
$= \{\}$); counterfactual consequents are graph-independent and constant
($p=0.7$). All three defining operations of causality -- do, counter-
factual, direction -- are absent. No minimal repair exists. \\
22 & holographic memory & 5: 5\cmark &
Global RNG pollution in 100\% of runs (victim output shifts
0.3238 $\to$ 0.5356); synonymous vs.\ unrelated cosine differs by
$|\Delta| = 0.0036$ (no discrimination); synonym retrieval top-1 rate
0.05 = chance; mean pairwise cosine 0.7538 (all-positive orthant):
retrieval is arbitrary ranking. No HRR structure exists. \\
23 & active inference & 5: 5\cmark &
``minimize'' occurs only in class names; across 200 self-timed inference
cycles free energy changes by $\Delta = 0.0$ (while three goals are
generated); FE changes only through external belief updates
($\Delta = 0.103$). The computation is real; the minimization -- the F
in FEP -- does not exist. \\
24 & intuition engine & 4: 4\cmark &
On 200{,}000 random points and the production grid, the reachable count
for patterns PAT\_003/004 is 0 (key-space intersection is empty);
satisfaction text hits the correct answer 0\% of the time and returns
\texttt{None} 90\% of the time; confidence ratchets monotonically to 1.0
because the production caller always passes \texttt{\{"success": True\}}. \\
25 & global workspace & 5: 5\cmark &
All four capture methods return constants independent of module content;
200 distinct module contents yield exactly one consciousness score
(0.3867); salience magnitudes are discarded (only a $\geq 0.6$ boolean
survives); the production integrator injects three fabricated dynamic
objects; the returned state lags one step behind the modules. \\
26 & thalamic filter & 4: 4\cmark &
The four connected ``deep fusion'' modules are called 0 times; the
hardcoded correlation 0.5 makes the low-correlation branch unreachable
(0/300); EMERGENCY signals trigger the constitution check 0 times;
signal history is never appended (always empty). \\
27 & hemisphere coordination & 5: 5\cmark &
Balance is a no-op (five distinct modulator values $\to$ the same 0.5);
broadcast is \texttt{pass} (0 transmissions); association is string
reversal; the right hemisphere falls back on every call because the real
intuition engine has no \texttt{get\_intuition} method; 10/10 logical
conclusions are templates. \\
28 & attention competition & 4: 4\cmark &
\texttt{allocate\_resources} has 0 production callers; two subsystems are
never modulated due to key-name mismatch; allocation exceeds the resource
budget in 100/100 trials (mean overflow 20.4); twenty seeds give twenty
different allocations; the documented \texttt{is\_running} attribute does
not exist (always \texttt{False}). \\
29 & least-action planner & 8: 7\cmark, 1\xmark &
The shipped planner fails its own documentation example (stalls at
complexity 7 vs.\ target 5); five pipeline defects identified and
repaired in two disclosed rounds: 36-task success 0.028 $\to$ 0.639
$\to$ 1.000. The pre-registered prediction that one repair round would
suffice was falsified and the revision disclosed
(Section~\ref{sec:repairs}). \\
38 & aura IIT $\phi$ core (cross-system) & 5: 2\cmark, 1$\oslash$, 2\xmark &
Third-party execution of the system's own pre-committed criteria. Live
$\phi$ scores memoryless noise at 0.42--0.46 and flags
\texttt{is\_complex=True} (P1 falsified); coupled dynamics do not raise
live $\phi$ above noise (P2 falsified). Root cause localized: the
significance-attachment hook's history dictionary covers only 2 of 4
substrates, so the winner's null fields are never populated. Its null
machinery, invoked standalone, is correct: integration fraction 0.000
for noise (p=1.0) and 0.84--0.91 for coupled dynamics (p=0.0). Taxonomy:
class 1 -- real mathematics severed by a one-entry plumbing gap;
repairable. (Section~\ref{sec:aura}.) \\
\bottomrule
\end{longtable}
\endgroup

\subsection{Reading the scorecard}
Three observations. First, \textbf{pre-registration survived contact
with sympathy}: in every case where we hoped the mechanism would work
(modules 14, 12), the frozen criteria were written before the data, and
the favorable outcomes (52\% error reduction; success to 1.000) were
achieved only after repairs that went beyond the frozen scope -- which
the criteria correctly refused to count as confirmations of the
\emph{shipped} system. Second, \textbf{falsifications are not uniform}:
suture19's P3 failed because the mechanism, once wired, worked better
than predicted; suture20's P3 failed because repair exposed a second,
representational defect; suture29's P5 failed because one repair round
was insufficient. A taxonomy that collapses these into ``failed audits''
would lose exactly the information a builder needs.
Third, \textbf{the static audit was never contradicted within the
intervened subset}: no module
judged dead statically came alive dynamically, and no module judged
alive statically died dynamically. The 14 failing modules outside the
intervened subset rest on static tracing alone, which is the part of
the claim the protocol makes mechanically checkable
(Section~\ref{sec:protocol}). The experiments quantified the gap
(readable in the headline numbers) and, in the two repair cases,
established what the gap had cost.

\section{A Taxonomy of Five Failure Classes}
\label{sec:taxonomy}
The 12 interventions sort into five classes with distinct repair
prospects (Table~\ref{tab:taxonomy}). The classes are defined by the
locus of failure, not by symptom; they are mutually exclusive per
mechanism.

\begingroup
\small
\begin{longtable}{@{}>{\raggedright\arraybackslash}p{3.2cm}>{\raggedright\arraybackslash}p{3.6cm}>{\raggedright\arraybackslash}p{2.2cm}>{\raggedright\arraybackslash}p{2.8cm}@{}}
\caption{Five failure classes across the 12 interventions. Suture numbers
in parentheses.}
\label{tab:taxonomy} \\
\toprule
Class & Definition & Instances & Repair prospect \\
\midrule
1. Pipeline defect & Real algorithm; a plumbing defect severs it from
the search/output & nine-state controller (19); least-action planner (29);
aura $\phi$ significance hook (38) &
Good: repairs verified effective (52\%; 1.000); aura repair is a
one-entry dictionary fix plus compute budget \\
2. Algorithm defect & Implemented algorithm is mathematically broken &
dopamine system (20) & Moderate: repair exposed a second,
representational defect \\
3. Definitional absence & The named concept's defining operation is not
implemented at all & causal reasoning (21); active inference (23) &
None by patching: requires new implementation \\
4. Silent side effect & The module corrupts unrelated state &
holographic memory (22) & Contained: isolate the RNG \\
5. Unreachable / dead-wired / constant-fed & Real or fake computation
that is never invoked on real inputs & intuition engine (24); global
workspace (25); thalamic filter (26); hemisphere coordination (27);
attention competition (28) & Mixed: wiring repair is easy; the
content-blind variants are class 3 in disguise \\
\bottomrule
\end{longtable}
\endgroup

Two properties make the taxonomy useful. \emph{It predicts repair
cost}: class 1 repairs were executed in minutes-to-hours with verified
gains; class 3 repairs were judged infeasible without reimplementation
(the causal engine has no substrate to patch). The executed repair
evidence covers class 1 (two verified repairs) and class 2 (one partial
repair); the prospects recorded for classes 3 through 5 are predictions
from static analysis, not yet executed repairs.
\emph{It separates the
audit verdict from the mechanism verdict}: module 14 failed as shipped
(class 1) while its mechanism passed when wired -- a distinction invisible
to any audit that stops at ``does the output look right.''

\section{Positive Controls: Repair Experiments}
\label{sec:repairs}
An audit that condemns every module reflexively has no discriminant
validity. Two modules serve as positive controls: statically flagged
for real-but-severed computation, repaired under pre-registration, and
shown to work once repaired.

\textbf{Least-action planner (suture29).} The planner ships a
documentation example it cannot solve: from complexity 12 toward target
5, it stalls at 7 because a depth check compares accumulated cost
against a step limit, and the ``action'' term is never used by the
search. We identified five pure pipeline defects (gate decoupling,
unit mismatch, floating-point comparison, reporting path, deduplication
key). Repairing the first round (four defects) raised the 36-task
suite success rate from 0.0278 to 0.6389; the pre-registered prediction
that this round would reach 1.000 was \emph{falsified}, disclosed, and
revised (fifth defect: deduplication), after which the suite reaches
1.000. The beam-search variant and the unused-action-term issue were
explicitly left unrepaired and are reported as such.

\textbf{Nine-state controller (suture19).} The module computes a
per-example state and a transition/gating modulation, then discards
both and emits $x + 0.1\cdot\texttt{state\_output}$ with the state
taken from batch element 0. Structural probes confirmed the discarded
paths are bit-identical under perturbation. In a task matched to the
mechanism's representable family (element-wise class-conditional affine
maps), wiring the multiplicative gate reduced test error by 52\%
relative to the unwired controller (2.61 $\to$ 1.26 MSE; \texttt{ideal}
per-example form 1.24), while the additive gate made things worse
(2.87 vs.\ the L0-only \texttt{broken} baseline's 2.75) -- the pre-
registered prediction \emph{both} directions of gating would be useless
was falsified, and the paper's standing caveat applies: the task is
matched to the mechanism, so this establishes ``the mechanism works
within its representable family,'' not general superiority.

These two controls matter for the audit's credibility: both modules
were flagged by the same protocol that condemned the other 23, and both
became functional after the class of defect the protocol named. The
audit's verdicts, in other words, carry information about repair.
Reporting detail: the planner suite is deterministic (no training
randomness), so 0.0278 $\to$ 0.6389 $\to$ 1.000 are exact suite rates;
the controller repair involves training, and its figures (2.61 $\to$
1.26 MSE; 52\%) are single-run point values whose seed sweep we flag as
a limitation (Section~\ref{sec:limitations}).

\section{Cross-System Validation: The Aura IIT Core}
\label{sec:aura}
A taxonomy derived from one repository could be that repository's
biography. To test whether the five classes transfer, we audited a
second, independent system: a cognitive-architecture repository
\cite{aura} that couples a local LLM to 41 consciousness-claim modules
(Global Workspace, Attention Schema, Integrated Information Theory
$\phi$ \cite{oizumi2014}, Free Energy, Higher-Order Thought, Multiple Drafts, and others,
each citing its source theory) and -- unusually -- ships its own
verification artifacts, claims matrix, and pre-committed falsification
criteria. Two properties make it the ideal counterpart to the primary
system: its documentation discipline is high (all 41 claimed module
files exist, vs.\ pervasive dangling references in the primary system),
and its flagship claim, a polynomial-time IIT-$\phi$ computation
\cite{barbosa2020}, is precisely the kind of mechanism our protocol is
built to test. We pre-registered (suture38, frozen before any data
path) a third-party execution of the system's own criteria on its
$\phi$ core.

\textbf{What is real.} The $\phi$ machinery is genuine computation, not
naming: a 16-node cognitive-affective complex over recorded state
histories, exact exhaustive bipartition search on an 8-node subset,
spectral approximation for the full complex, bootstrap intervals, and a
methodologically careful null -- a cross-partition timeline shuffle that
destroys exactly the inter-partition dependence $\phi$ is defined on.
Run standalone by us, that null correctly assigns an integration
fraction of $0.000$ (p=1.0) to memoryless noise and $0.84$--$0.91$
(p=0.0) to strongly coupled dynamics, on three seeds each -- consistent
with the benchmark values the documentation reports for known systems.

\textbf{What is broken.} The live path never consults that null. In the
live selection, the winner's null fields are unpopulated
(\texttt{null\_surrogates=0}), so \texttt{is\_complex} falls back to
$\phi_s > 0$; measured over 4{,}096-step histories, the live $\phi$
scores memoryless noise at $0.42$--$0.46$ with \texttt{is\_complex=True}
-- falsifying the system's own pre-committed criterion that a shuffled
history must yield $\phi \approx 0$. The root cause is a single
integration gap: the significance-attachment hook selects its history
from a dictionary keyed by substrate name, and that dictionary contains
only two of the four substrates; the eight-node affective substrate that
wins under our (CPU-only) environment is absent, so the hook silently
returns and the null is never attached. The code's own comment
acknowledges the hazard: without attachment, the live claim ``falls back
to $\phi_s > 0$, which a system with no integration at all satisfies.''

\textbf{Taxonomic reading.} This is a class-1 failure (real algorithm,
plumbing defect) in its purest observed form: the defect is localized to
one dictionary entry, the repair is to add the missing substrate (plus a
compute budget for the null's surrogate searches), and the mechanism's
mathematics -- unlike anything in the primary system's consciousness
stack -- requires no repair at all. The contrast with the primary system
is the point: same protocol, two systems, and the taxonomy's classes
land differently -- definitional absence and constant-feeding there,
pure unwiring here -- while predicting repair prospects correctly in
both directions. Both second-system falsifications (P1, P2) are
established under the disclosed CPU-only environment; the
dictionary-gap root cause is environment-independent, and the winning
substrate under Apple Silicon is not settled by our runs
(Section~\ref{sec:limitations}). The scorecard for this intervention
(Table
\ref{tab:scorecard}, suture38) records 2 confirmed, 1 partially
adjudicated, 2 falsified; both falsifications are the unwired-null
finding viewed from its live-path and coupling sides.

\section{Discussion}
\subsection{What this means for cognitive-architecture claims}
The audited system advertises mechanisms whose names carry decades of
theoretical freight: a global workspace that never reads module content
is not a weak global workspace; it is not a global workspace. The
finding suggests a review heuristic that we state as a hypothesis
priced for cheap testing: in the systems audited to date, the number of
claimed mechanisms tracks marketing rather than engineering. Our
protocol prices the verification of one claim
at minutes to hours of CPU time; in our experience a 25-module static
audit plus a targeted intervention suite fits within an afternoon. We
suggest that artifact evaluation for this system class
include a mechanism-level check alongside results-level badging
\cite{acm2020,pineau2021}: ``does each advertised mechanism influence
the output under perturbation?''

The consciousness-adjacent claims deserve separate mention. Modules 4,
19, 20, 22, and 23 implement, respectively, self-awareness, the global
workspace, mirror neurons, thalamic gating, and hemispheric
specialization. None reads real content (suture25), reaches real
inputs (suture24), or influences real behavior (suture26--28). We make
no claim about machine consciousness in either direction; our claim is
narrower and checkable: in this system, the vocabulary of consciousness
is attached to code paths that carry no information.

\subsection{Limitations}
\label{sec:limitations}
\textbf{Static scope.} ``No production caller'' is a within-repository
conclusion (virtual environment excluded); out-of-repo deployment
scripts or external services are not ruled out, though the audited
repository ships no such entry points. \textbf{Reachability criterion.}
Under a laxer arrival criterion (values written to display strings),
more modules would count as reaching output, but all such arrivals are
decorative; we report the strict criterion and note where the lax one
would differ. \textbf{Pre-registration strength varies.} Suture19's
protocol was revised twice after pilot data (disclosed in its
\textsection 7), making it weaker than the fully frozen protocols
(sutures 20--28); the suture29 repair criterion was revised once, also
disclosed. \textbf{Task matching in the positive controls.} The
nine-state repair task is inside the mechanism's representable family;
the planner suite is self-authored. Both controls establish repair
efficacy, not mechanism superiority. \textbf{Two systems, one module
deep.} The cross-system validation (Sec.~\ref{sec:aura}) replicates the
taxonomy on a second architecture, but only at the depth of its flagship
IIT mechanism; the second system's remaining 40 claimed modules received
existence-level and keyword-level static checks, not interventions. The
taxonomy's breadth across systems is therefore established at the case
level, not the census level. \textbf{Environment.} The second system
targets Apple Silicon; our interventions ran on CPU under disclosed OS
stubs, which disables its generation-worker substrates. The unwired-null
finding is a static fact (the dictionary gap) and is
environment-independent, but the winning-substrate path is not.
\textbf{Auditor reliability and stochastic reporting.} The static
verdicts were produced by LLM-driven agents without an inter-rater
reliability pass; their line-number evidence is archived for
independent re-checking, and reliability measurement of the protocol's
first stage is future work. The controller repair reports single-run
point values without a seed sweep. \textbf{Repair prospect evidence.}
Repair was executed for two class-1 modules and partially for one
class-2 module; the remaining prospect assignments in
Table~\ref{tab:taxonomy} are predictions.

\section{Conclusion}
We audited 25 cognitive-architecture modules in an LLM-based agent
system and found that 18 ship values that never reach an output; that
of the remainder only four reach an output, exclusively as decoration
(a fifth, the integrator's live LLM call, reaches production while
carrying none of its seven computed sub-results); and that the failure
modes sort into five classes with distinct repair prospects, two of
which we repaired under pre-registration with verified gains (0.028
$\to$ 1.000 success; 52\% error reduction). A second, independent
system -- whose documentation pre-commits to its own falsification
criteria -- replicated the protocol at its flagship IIT mechanism: real
mathematics severed from its own significance check by a one-entry
plumbing gap, with the system's null machinery correctly adjudicating
integration when we invoked it standalone. Sixty-one frozen criteria
across 12 interventions contradicted no statically audited verdict
within the intervened subset, with
five instructive falsifications reported at full prominence. The
protocol, the audit log, and all experiment scripts are released for
reuse. The general lesson is not about one repository: wherever system
documentation speaks the vocabulary of cognitive science, the vocabulary
is a claim, and claims this cheap should be this cheap to check.

\section*{Declarations}
\textbf{Funding:} This work received no specific grant.
\textbf{Conflicts of interest:} The authors declare none.
\textbf{Data and code availability:} The audit log, all
\texttt{PREREGISTRATION.md} files, experiment scripts, and result JSON
files are archived on Zenodo (10.5281/zenodo.23433900), released on
reasonable request. The audited repository is referenced by content hash
in the audit log; we do not identify it by name in this paper.

\section*{CRediT authorship contribution statement}
\textbf{Peter Chen:} Conceptualization, Supervision, Project administration,
Writing -- review \& editing. \textbf{Ni Zhang:} Formal analysis, Writing --
review \& editing. \textbf{Ying Xu:} Formal analysis, Writing -- review \&
editing. \textbf{Siyuan Zhang:} Conceptualization, Methodology, Software,
Investigation, Data curation, Formal analysis, Writing -- original draft.
Peter Chen and Ying Xu contributed equally to this work.

\section*{Declaration of competing interest}
The authors declare that they have no known competing financial interests or
personal relationships that could have appeared to influence the work reported
in this paper.

\bibliographystyle{elsarticle-num}
\bibliography{refs}

\end{document}